\documentclass[]{bytedance_seed}

\usepackage[toc,page,header]{appendix}

\usepackage{minitoc}

\usepackage{booktabs} 
\usepackage{pifont}   

\newcommand{\cmark}{\ding{51}} 
\newcommand{\xmark}{\ding{55}}

\title{ FlowAct-R1: Towards Interactive Humanoid Video Generation }

\author[*]{Lizhen Wang}
\author[*]{Yongming Zhu}
\author[*]{Zhipeng Ge}
\author[*]{Youwei Zheng}
\author[*]{Longhao Zhang}
\author[*\dagger]{Tianshu Hu}
\author{Shiyang Qin}
\author{Mingshuang Luo}
\author{Jiaxu Zhang}
\author{Xin Chen}
\author{Yulong Wang}
\author{Zerong Zheng}
\author{Jianwen Jiang}
\author{Chao Liang}
\author{Weifeng Chen}
\author{Xing Wang}
\author{Yuan Zhang}
\author{Mingyuan Gao}

\affiliation{ByteDance Intelligent Creation}

\contribution[*]{Core Contributors}
\contribution[\dagger]{Corresponding author}

\abstract{
Interactive humanoid video generation aims to synthesize lifelike visual agents that can engage with humans through continuous and responsive video.
Despite recent advances in video synthesis, existing methods often grapple with the trade-off between high-fidelity synthesis and real-time interaction requirements.
In this paper, we propose \textbf{FlowAct-R1}, a framework specifically designed for real-time interactive humanoid video generation.
Built upon a MMDiT architecture, FlowAct-R1 enables the streaming synthesis of video with arbitrary durations while maintaining low-latency responsiveness.
We introduce a chunkwise diffusion forcing strategy, complemented by a novel self-forcing variant, to alleviate error accumulation and ensure long-term temporal consistency during continuous interaction.
By leveraging efficient distillation and system-level optimizations, our framework achieves a stable 25fps at 480p resolution with a time-to-first-frame (TTFF) of only around 1.5 seconds. 
The proposed method provides holistic and fine-grained full-body control, enabling the agent to transition naturally between diverse behavioral states in interactive scenarios.
Experimental results demonstrate that FlowAct-R1 achieves exceptional behavioral vividness and perceptual realism, while maintaining robust generalization across diverse character styles.
}

\date{\today}
\correspondence{Tianshu Hu at \email{tianshu.hu@bytedance.com}}

\checkdata[Project Page]{\url{https://grisoon.github.io/FlowAct-R1/}}

\begin{document}
\maketitle


\vspace{-1.0em}

\begin{figure*}
  \centering
  \includegraphics[width=0.92\textwidth]{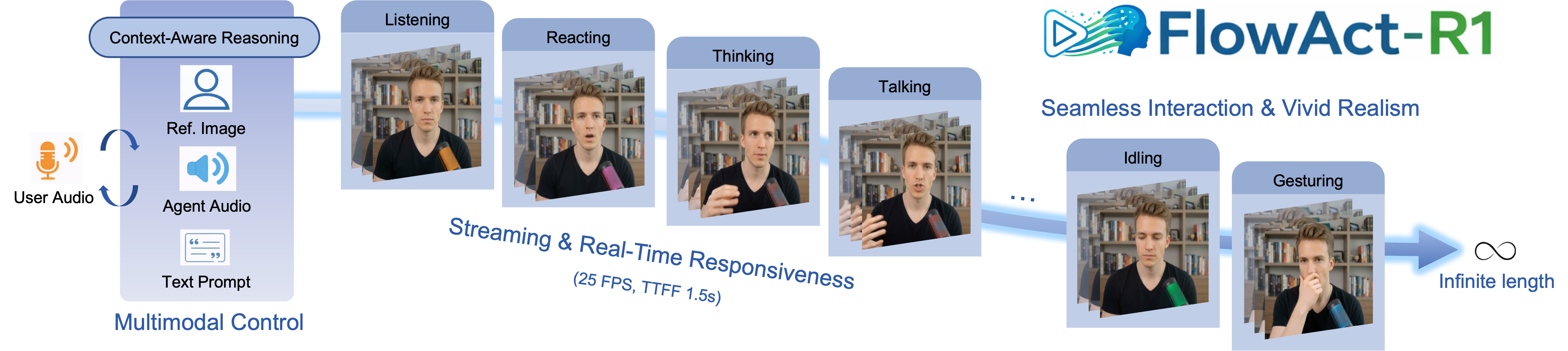}
\caption{We present FlowAct-R1, a novel framework that enables lifelike, responsive, and high-fidelity humanoid video generation for seamless real-time interaction.}
\label{fig:teaser}
\end{figure*}

\section{Introduction}

Enabling visual humanoid agents to engage in real-time, natural interactions with humans is a long-standing objective in the research community \cite{low2025talkingmachinesrealtimeaudiodrivenfacetimestyle, ao2024bodyofher, huang2025liveavatarstreamingrealtime, zhu2024infp, guo2025arigautoregressiveinteractivehead, chen2025midasmultimodalinteractivedigitalhuman, agrawal2025seamlessinteractiondyadicaudiovisual, xie2025xstreamerunifiedhumanworld}. In the task of interactive humanoid video generation, the model is required to synthesize naturalistic videos conditioned on conversational contexts (e.g., audio and text) from both the user and the agent.
To realize this vision, several critical challenges must be addressed. First, the model must support streaming and real-time video generation to ensure low-latency and responsive interaction \cite{zhen2025tellerrealtimestreamingaudiodriven, yin2025causvid, huang2025selfforcing, chen2024diffusionforcingnexttokenprediction, shin2025motionstreamrealtimevideogeneration, kodaira2025streamditrealtimestreamingtexttovideo, liu2025rollingforcingautoregressivelong}. Furthermore, as multi-round interaction inherently involves extended durations, maintaining visual quality and temporal consistency in long-form video remains a non-trivial task \cite{yi2025magic, yang2025longlive, li2025stablevideoinfinityinfinitelength, oshima2025worldpackcompressedmemoryimproves, zhang2025framepack, yu2025contextmemorysceneconsistentinteractive, deepmind2024genie3}. Finally, humanoid interaction involves a variety of behavioral states such as speaking, listening, reflecting, and idling. The ability to seamlessly transition between these dynamic states while producing plausible behaviors is essential for achieving perceptual realism and lifelike engagement \cite{low2025talkingmachinesrealtimeaudiodrivenfacetimestyle, ao2024bodyofher, zhu2024infp, guo2025arigautoregressiveinteractivehead, agrawal2025seamlessinteractiondyadicaudiovisual}.
To this end, we introduce \textbf{FlowAct-R1}, a framework specifically designed for interactive humanoid generation. Built upon a MMDiT architecture \cite{gao2025seedance10exploringboundaries, seedance2025seedance15pronative}, our approach enables the streaming synthesis of video with arbitrary lengths. It achieves real-time performance while maintaining low-latency responsiveness. The framework facilitates fine-grained controllability over the generated humanoid video—encompassing lip-sync, facial expressions, body gestures, and object interactions—allowing it to adapt naturally to various behavioral states during interactions. Our method produces lifelike videos and demonstrates robust generalization across diverse characters. We believe this framework paves the way for applications such as live streaming, virtual companionship, and video conferencing.

Early research on humanoid video generation has primarily centered on lip-synchronization \cite{zhou2021pose, 10.1145/3394171.3413532, Zhong_2023_CVPR, guan2023stylesync, zhang2024personatalk, thies2020nvp}. By conditioning on audio signals, these methods synthesize mouth movements precisely aligned with speech. While these techniques have reached commercial maturity \cite{heygen2024avatar, synthesia2024digitalhuman}, their scope remains largely confined to the facial region, lacking fine-grained control over full-body gestures. This inherent limitation hinders the generation of highly expressive and naturalistic behaviors necessary for truly lifelike interaction.
Recently, diffusion-based generative models have demonstrated significant breakthroughs in visual quality \cite{jiang2025omnihuman15instillingactivemind, klingteam2025klingavatar20technicalreport, gan2025omniavatarefficientaudiodrivenavatar, kong2025let, cui2024hallo3}. Although these approaches can precisely manipulate body dynamics to synthesize vivid motion, they often suffer from heavy computational overhead, leading to prohibitively slow inference speeds. Furthermore, most existing diffusion frameworks are restricted to short-clip generation and lack support for continuous streaming, which limits their deployment in real-time interactive scenarios.
Meanwhile, several efforts have specifically targeted interactive tasks. While methods such as INFP \cite{zhu2024infp} and ARIG \cite{guo2025arigautoregressiveinteractivehead} enable real-time streaming for long-form video, they are predominantly constrained to portrait-style cropping. Other approaches, such as TalkingMachines \cite{low2025talkingmachinesrealtimeaudiodrivenfacetimestyle} and LiveAvatar \cite{huang2025liveavatarstreamingrealtime}, achieve real-time streaming performance through model distillation or engineering optimizations but still exhibit a perceptual gap in terms of behavioral vividness and naturalness.

\begin{table}[t]
\centering
\begin{tabular}{l|ccccc}
\toprule
\multicolumn{1}{c|}{\textbf{Method}} & Stream & Real-Time & Full-body Control & Generalization & Vividness \\ 
\midrule
\quad Neural Voice Puppetry \cite{thies2020nvp} & \cmark & \cmark & \xmark & \xmark & \xmark \\
\quad INFP \cite{zhu2024infp} & \cmark & \cmark & \xmark & \cmark & \cmark \\
\quad Omnihuman-1.5 \cite{jiang2025omnihuman15instillingactivemind} & \xmark & \xmark & \cmark & \cmark & \cmark \\
\quad KlingAvatar 2.0 \cite{klingteam2025klingavatar20technicalreport} & \xmark & \xmark & \cmark & \cmark & \cmark \\
\quad LiveAvatar \cite{huang2025liveavatarstreamingrealtime} & \cmark & \cmark & \cmark & \cmark & \xmark \\
\midrule
\quad FlowAct-R1 (ours) & \cmark & \cmark & \cmark & \cmark & \cmark \\
\bottomrule
\end{tabular}

\caption{Comparison of state-of-the-art humanoid video generation methods. FlowAct-R1 simultaneously achieves streaming, real-time generation with fully controllable, generalization, and lifelike video generation capacity.}
\label{tab:comparison}
\end{table}

In light of the limitations identified above, we introduce \textbf{FlowAct-R1}, a framework designed expressly for interactive humanoid video generation. A comprehensive comparison between our method and existing state-of-the-art approaches is presented in Table \ref{tab:comparison}, highlighting FlowAct-R1's unique capability to simultaneously achieve streaming, real-time performance, and high-fidelity behavioral expressivity. Our main contributions are summarized as follows:
\begin{itemize}
    \item \textbf{Streaming and Infinite-Length Generation}: Leveraging a chunkwise diffusion forcing strategy, we adapt a MMDiT backbone \cite{gao2025seedance10exploringboundaries, seedance2025seedance15pronative} for streaming synthesis. To alleviate error accumulation over extended sequences, we design a self-forcing \cite{huang2025selfforcing} variant to bridge the gap between training and inference, complemented by a simple memory strategy to promote long-term temporal consistency.
    \item \textbf{Real-time Inference with Low Latency}: We optimize the model across both algorithmic and system-level dimensions. By employing efficient distillation techniques \cite{yin2024onestep, yin2024improved, salimans2022progressivedistillationfastsampling, song2023consistencymodels}, we reduce the denoising process to only 3 NFEs. Combined with operator-level optimizations and parallel computing, our framework achieves real-time 480p video generation at 25fps with a time-to-first-frame (TTFF) of around 1.5s.
    \item \textbf{Vividness and Generalization}: Our method preserves the robust generalization of its foundational model, enabling high-fidelity synthesis from a single reference image across various character styles. Furthermore, it supports holistic control via audio and text, allowing for natural transitions between diverse interactive states and significantly outperforming existing SOTA models in behavioral vividness.
\end{itemize}

\section{Approach}
\label{sec:method}
We propose \textbf{FlowAct-R1}, a real-time streaming video generation framework architected upon \textit{Seedance}\cite{gao2025seedance10exploringboundaries, seedance2025seedance15pronative}. Serving as the backbone, Seedance is a Multimodal Diffusion Transformer (MMDiT) featuring native cross-modal alignment capabilities. To achieve indefinite-length streaming, FlowAct-R1 employs a chunkwise diffusion forcing strategy~\cite{kodaira2025streamditrealtimestreamingtexttovideo, chen2024diffusionforcingnexttokenprediction, huang2025selfforcing, liu2025rollingforcingautoregressivelong, yang2025longlive} augmented by a structured memory bank. By integrating short-clip curriculum training with system-level inference optimizations, we realize low-latency, interactive video generation. Furthermore, through fine-grained annotation on behavior-rich human datasets, our method enables the generation of vivid, text-controllable human dynamics.

\begin{figure*}[h]
  \centering
  \includegraphics[width=1.0\textwidth]{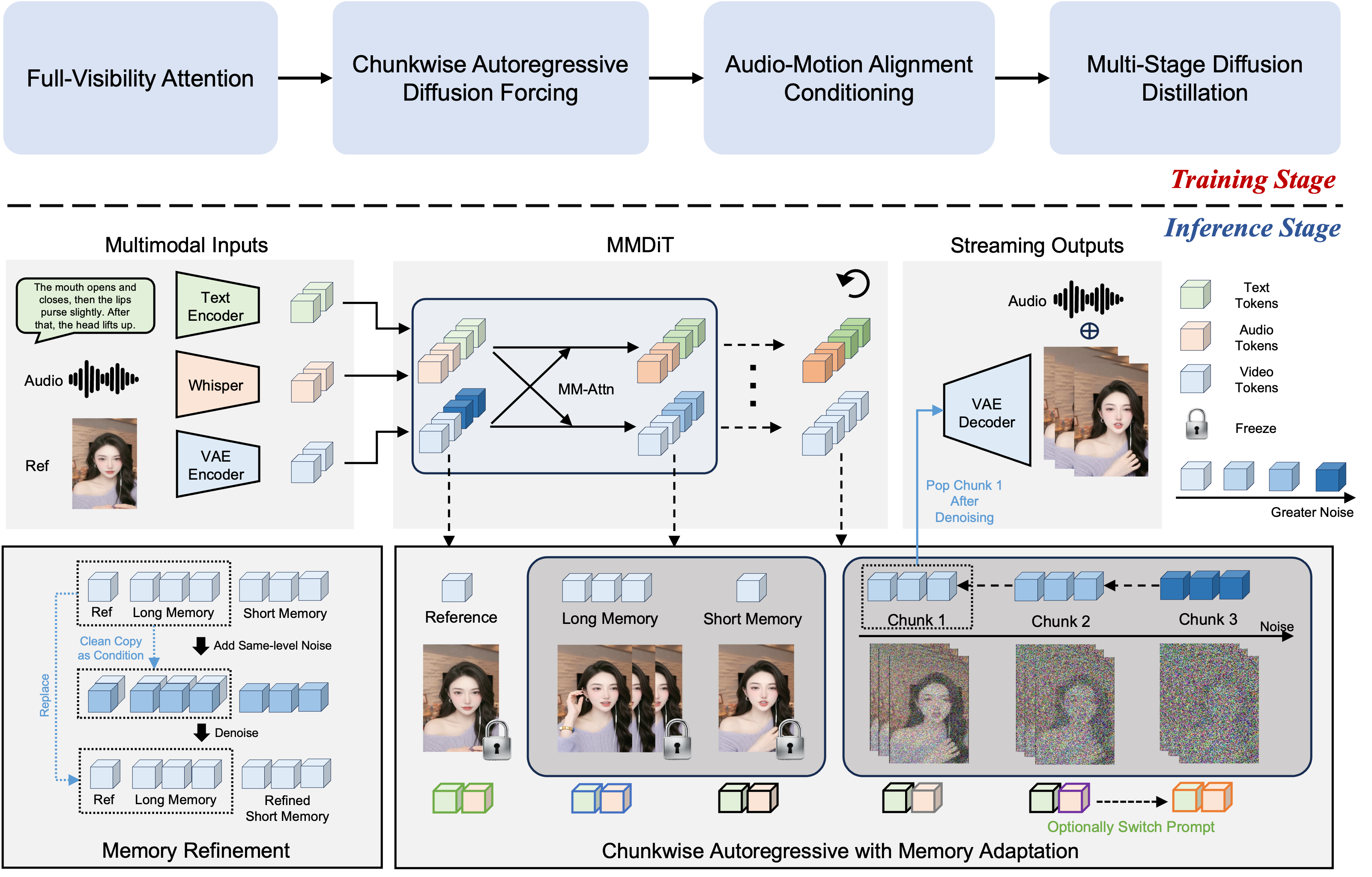}
\caption{Overview of the FlowAct-R1 framework. It consists of training and inference stages: training integrates converting base full-attention DiT to streaming AR model via autoregressive adaptation, joint audio-motion finetuning for better lip-sync and body motion, multi-stage diffusion distillation; inference adopts a structured memory bank (Reference/Long/Short-term Memory, Denoising Stream) with chunkwise autoregressive generation and memory refinement. Complemented by system-level optimizations, it achieves 25fps real-time 480p video generation (TTFF ~1.5s) with vivid behavioral transitions.}
\label{fig:overview}
\end{figure*}

\subsection{Overall Pipeline}
\label{sec:pipeline}

FlowAct-R1 inherits the foundational architecture of Seedance. The input video stream is compressed temporally and spatially into latent tokens via a VAE, while text prompts are encoded into semantic tokens. For the audio branch, inspired by OmniHuman-1.5~\cite{jiang2025omnihuman15instillingactivemind}, we utilize Whisper \cite{radford2022robustspeechrecognitionlargescale} to compress audio input into aligned acoustic tokens. These multimodal representations are fused within the MMDiT, where information extraction and exchange occur via cross-attention mechanisms. The efficient design of Seedance—characterized by reduced parameters, shot-based temporal slicing, and window-based spatial attention—provides the requisite computational speed for real-time interaction.

The core of our streaming inference is a fixed-size stream buffer tailored to the MMDiT's sequence length constraints. It comprises four key components:
\begin{enumerate}
    \item \textbf{Reference Latent:} A single input reference image used to anchor identity and temporal consistency.
    \item \textbf{Long-term Memory Queue:} A buffer of fully denoised latents (max size 3) from earlier chunks, preserving long-range action dependencies.
    \item \textbf{Short-term Memory Latent:} The most recently fully denoised latent, utilized to enforce local motion smoothness.
    \item \textbf{Denoising Stream:} A queue organized as 3 chunks $\times$ 3 latents per chunk, currently undergoing parallel gradient updates.
\end{enumerate}
This structured design guarantees a continuous output of 0.5 seconds of video for every 0.5 seconds of wall-clock time (corresponding to one chunk), supporting a 3-NFE (Number of Function Evaluations) inference cycle while maintaining temporal coherence.

\subsection{Training Procedure and Framework Design}
\label{sec:training}

Our training pipeline follows a three-stage curriculum: (1) Autoregressive Adaptation, transforming the general-purpose MMDiT into a streaming-compatible variant; (2) Joint Audio-Motion Training, integrating the speech module; and (3) Distillation, employing distillation to compress the inference process to 3 NFEs. We utilize a composite dataset of general video samples and conversation videos. All data is pre-processed into clips with dense annotations to facilitate granular streaming control.

To ensure robustness in streaming, we adopt two distinct AR training paradigms: intra-prompt segment training to learn local dependencies, and cross-prompt segment training to model smooth transitions between segments with differing text prompts. Additionally, native Image-to-Video (I2V) capabilities are retained via weighted loss during training to guarantee coherent initialization of the first generated segment. The final distillation stage balances denoising quality with real-time performance without sacrificing core capabilities.

Furthermore, we implement a fake-causal attention mechanism during both training and inference. Specifically, the attention mask is designed such that the denoising stream has full visibility over the reference, memory, and itself, whereas the reference and memory components are restricted from attending to the denoising stream. This asymmetric design reduces computational overhead and ensures that the fully denoised information remains stable across deepening DiT layers, serving as a robust and uncorrupted anchor for the generation process.
Additionally, drawing on Self-Forcing++~\cite{cui2025self}, to align the cumulative memory errors between training and inference, we use an intermediate trained model to perform noise injection and denoising on groundtruth video latents to obtain generated-GT-latents. During training, we probabilistically select generated-GT-latents instead of GT-latents when sampling memory components to simulate inference-stage memory errors, which helps mitigate cumulative errors during inference.

\subsection{Multimodal Fusion}
\label{sec:fusion}

The MMDiT architecture of Seedance \cite{gao2025seedance10exploringboundaries, seedance2025seedance15pronative} inherently supports text-visual fusion. For the audio branch, inspired by OmniHuman-1.5~\cite{jiang2025omnihuman15instillingactivemind}, we adopt an IP-Adapter-style approach to correlate audio signals with fine-grained motions, such as lip synchronization and body dynamics, via cross-attention. 

During both training and inference, the video is processed at 25 FPS. Audio input (16kHz) is converted into 25 features per second using Whisper. These features are then aggregated into condition vectors with temporal overlap.
Text prompts are specifically annotated to describe detailed human behaviors within short-range intervals and periodically updated during inference. Utilizing such short, action-dense clips better adapts to low-latency streaming requirements while ensuring accurate motion response and behavioral diversity.

\subsection{Model Acceleration}
\label{sec:distillation}

We design a multi-stage distillation pipeline specifically tailored to FlowAct-R1’s architecture and its unique training-inference paradigm. This pipeline progressively reduces the sampling cost to a highly efficient 3 NFEs (chunk-size=3, micro-step=1, without CFG), achieving an $8\times$ acceleration while maintaining competitive synthesis quality.

Prior to step distillation, we first eliminate the overhead of classifier-free guidance (CFG). This is achieved by injecting an auxiliary CFG embedding layer and distilling outputs from various guidance scales into a single, unified model. 
We then perform naive step distillation \cite{salimans2022progressivedistillationfastsampling}, where original NFEs are partitioned into three segments. Within each segment, constituent micro-steps are distilled into a single step.
Following step distillation, we apply few-step score distillation - DMD \cite{yin2024improved}. Crucially, both the student and fake models are initialized from the checkpoint obtained in the previous stage to ensure training stability and convergence.
To better align the distilled model with streaming generation trajectories, we further modify DMD by chunking training videos and explicitly simulating FlowAct-R1’s progressive rollout behavior during online generation and backward simulation.

Beyond diffusion distillation, we implement a comprehensive suite of optimizations to achieve consistent real-time generation performance.
Specifically, we strategically employ FP8 quantization across a selection of attention and linear layers, effectively boosting inference throughput with minimal impact on synthesis quality.
To further streamline distributed inference, we transition from token-level sequence parallelism to a frame-level hybrid-parallel strategy. This reconfiguration significantly reduces the all-to-all communication overhead and mitigates network bottlenecks. 
To minimize the data movement overhead between HBM and SRAM \cite{dao2022flashattention}, frequent operators are fused into single kernels within each DiT block.
Furthermore, we decouple DiT denoising and VAE decoding into an asynchronous pipeline to enable concurrent execution.

Consequently, these optimizations enable FlowAct-R1 to achieve real-time 480p video synthesis at 25 fps on the NVIDIA A100 platform, while restricting the Time-to-First-Frame (TTFF) to approximately 1.5 seconds.

\subsection{Inference Optimization}
\label{sec:optim}

To enhance consistency in long-duration videos and mitigate error accumulation during streaming, our empirical analysis reveals that the short-term memory exerts the most significant influence on the denoising stream, and consequently, cumulative errors manifest earliest within this component. To address this, we introduce a \textbf{Memory Refinement} strategy. 
At regular intervals, we conduct noise injection and denoising repair operations on short-term memory frames. During the denoising phase, a copy of reference and long-term memory are employed as stable guidance constraints.
This process effectively rectifies artifacts accumulated from continuous streaming. Combined with our structured memory bank and chunked denoising framework, this optimization ensures that our long-video generation maintains sustained motion smoothness and identity consistency.

\subsection{Multimodal Action Planning}
\label{sec:optim}

To enhance behavioral naturalness, we integrate a Multimodal Large Language Model (MLLM) into the action planning process. At regular short intervals, the latest audio segment (speech content) and reference image are fed into the MLLM, which predicts plausible subsequent actions aligned with contextual cues and visual constraints. These action priors guide the MMDiT backbone, ensuring smooth and natural transitions between interactive states.
\section{Experiments}

We compare FlowAct-R1 with three SOTA methods: KlingAvatar 2.0~\cite{klingteam2025klingavatar20technicalreport}, LiveAvatar~\cite{huang2025liveavatarstreamingrealtime}, and Omnihuman-1.5~\cite{jiang2025omnihuman15instillingactivemind}. Omnihuman-1.5 shares a similar network structure but only supports up to 30s video without real-time streaming; KlingAvatar 2.0 achieves 5min long-duration and high visual quality yet lacks streaming capability and suffers from motion repetition. LiveAvatar, based on Wan2.2-S2V-14B~\cite{gao2025wans2vaudiodrivencinematicvideo}, enables real-time streaming but also faces motion repetition issues that reduce naturalness.

\begin{figure*}[h]
  \centering
  \includegraphics[width=1.0\textwidth]{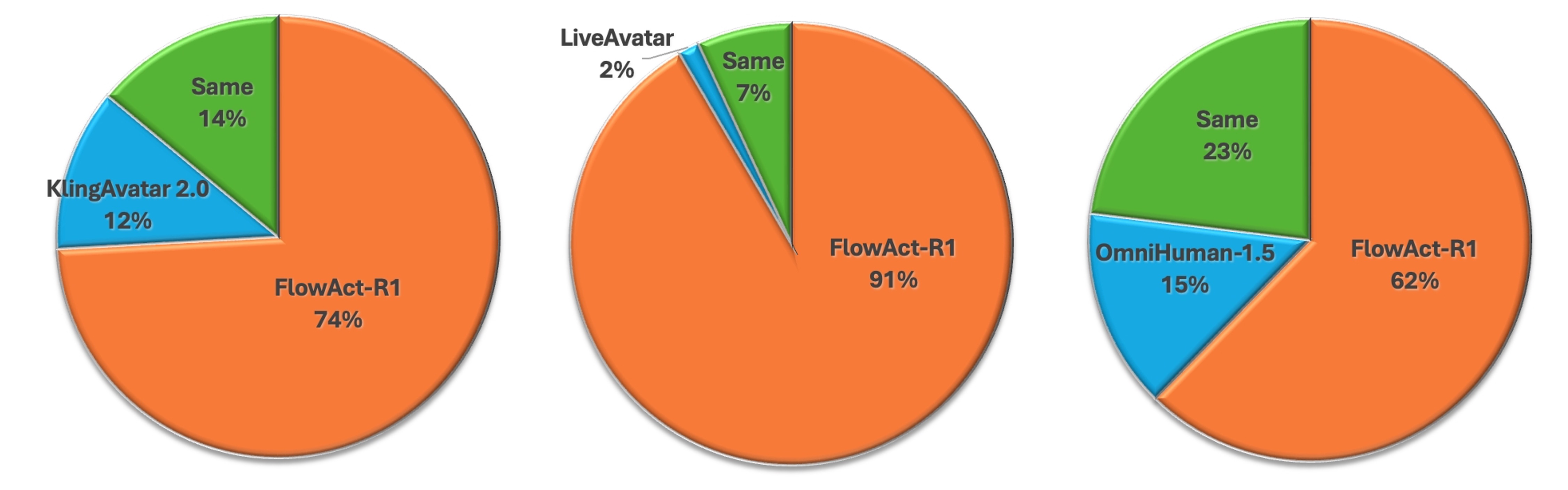}
\caption{Comparisons with KlingAvatar 2.0~\cite{klingteam2025klingavatar20technicalreport}, LiveAvatar~\cite{huang2025liveavatarstreamingrealtime}, and Omnihuman-1.5~\cite{jiang2025omnihuman15instillingactivemind} via a user study using the GSB (good-same-bad) metric. The $\textbf{\textcolor[RGB]{255,127,14}{orange segments}}$ indicate the percentage of user votes favoring FlowAct-R1 over other methods. Video demos are shown in our project page.}
\label{fig:exp}
\end{figure*}

We evaluate all methods via a user study using the GSB (good-same-bad) metric. The key evaluation metrics mainly include motion naturalness, lip-sync accuracy, frame structure stability, and motion richness. For consistency, FlowAct-R1 and LiveAvatar use full-length audio, while Omnihuman-1.5’s audio is truncated to 30s and KlingAvatar 2.0’s to 5min (matching their maximum video durations). We invited 20 participants for this user study. As shown in Fig.~\ref{fig:exp}, results show FlowAct-R1 outperforms competitors by simultaneously supporting long-duration streaming, real-time responsiveness (25fps at 480p, TTFF around 1.5s), and superior behavioral naturalness—attributed to MLLM-guided action planning and chunkwise diffusion forcing that mitigate motion repetition.

\section{Conclusion}

In this paper, we presented \textbf{FlowAct-R1}, an interactive humanoid video generation framework aimed at synthesizing lifelike agents that can engage with humans through continuous and responsive video.
Our framework enables streaming, arbitrary-duration video generation while maintaining superior temporal consistency and visual fidelity.
Through synergistic model-level distillation and system-level optimizations, our method achieves stable real-time performance with low-latency responsiveness.
Notably, the model delivers exceptional behavioral vividness and perceptual realism, capturing subtle human nuances for natural transitions across complex interactive states, while maintaining high-quality synthesis across diverse character styles from a single reference image.

\textbf{Ethical Considerations.} There is a potential risk that our method could be misused to fabricate deceptive or harmful content. To mitigate this, we are committed to the responsible deployment of our technology and will implement a rigorous access control policy for our core models, ensuring they are provided only to verified entities for legitimate and ethical use. Notably, all human images used in our demonstrations were generated by AI tools (e.g. Gemini or GPT-4o) to ensure privacy and copyright compliance.

\clearpage

\bibliographystyle{plainnat}
\bibliography{main}




\end{document}